# A Modification to Evidential Probability

Bülent Murtezaoğlu
mucit@cs.rochester.edu
Computer Science Department
University of Rochester
Rochester, NY 14627

Henry E. Kyburg
kyburg@cs.rochester.edu
Computer Science Department
University of Rochester
Rochester, NY 14627

## Abstract

Selecting the right reference class and the right interval when faced with conflicting candidates and no possibility of establishing subset style dominance has been a problem for Kyburg's Evidential Probability system. Various methods have been proposed by Loui and Kyburg to solve this problem in a way that is both intuitively appealing and justifiable within Kyburg's framework. The scheme proposed in this paper leads to stronger statistical assertions without sacrificing too much of the intuitive appeal of Kyburg's latest proposal.

## 1 Overview of the Problem

### 1.1 An Example

Let us consider a variant of the classic berries example.[1] Suppose a hungry agent has access to the following information:

- Between 70 and 90 percent of the red berries, sampled at some time in the past, were found to be good to eat.

- Between 30 and 50 percent of the berries picked on rainy days were found to be good to eat.

- Between 70 and 75 percent of berries from this region, sampled at some time in the past, were found to be good to eat.

- Between 35 and 45 percent of soft berries, sampled at some time in the past, were found to be good to eat.

- The berries at hand now are both red and soft and picked from this region and furthermore today is a rainy day.

---

[1] Due to Jerry Feldman.

The agent's problem is deciding whether or not it should eat the berries it has. This decision involves two distinct levels of analysis; the first one is deciding what indeed it can infer from its knowledge about berries in general about the particular berries it has, and the second one is, given its inferred knowledge about the berries it has, whether or not it should eat them.

It may be argued that de-coupling the inference and the decision procedures generally leads agents into lengthy computations even when the relevant utility values and practical concerns would dictate a certain decision, rendering the sophisticated inference procedure futile. In the example the agent may be making a choice between starvation and food poisoning and therefore the utilities involved with the choices would, for a sane agent, dictate that it should eat the berries regardless of what it can infer about their edibility. We will not, however, concern ourselves with such issues in this paper because the proposed method for this restricted case is computationally cheap.

In the example above, the agent has no knowledge of the subset relationships between the candidate reference classes. For instance, if it knew that the red berries that it has statistics about were in fact both red and soft, it could safely disregard the conflicting[2] statistics about soft berries [Kyburg, 1983]. Or if it had access to the joint information about red and soft berries found in this region on rainy days, it would not need to consider the conflicting statistics about the broader classes according to both Reichenbach and Kyburg. Our agent, however, does not have all the necessary bits of information conveniently available.

### 1.2 The General Case

Suppose we want to compute the probability of some object $o$ having a target property $T$, and we have knowledge about the classes $o$ belongs to and the in-terval valued measure of $T$ in those classes. More formally, our knowledge base contains the following statements:

- Sentences denoting set memberships
  "$x \in Y$"
  where $x$ is an object and $Y$ is a set.
- Sentences denoting subset relationships between classes
  "$Y \subset Z$"
  where $Y$ and $Z$ are sets.
- Sentences concerning proportions of sets of the form
  "$\%(T, S) = [p, q]$"
  where $T$ and $S$ are sets and $p$ and $q$ are some approximate representation of real numbers. These can be read as " the measure or the proportion of elements of set $S$ that have the property $T$ is in the interval $[p, q]$."

---

[2] *Conflict*, or *disagreement*, between two intervals $[p_1, q_1]$ and $[p_2, q_2]$ is defined as the case where neither $[p_1, q_1] \subseteq [p_2, q_2]$ nor $[p_2, q_2] \subseteq [p_1, q_1]$.

Using the above syntax and assumptions, we can state the general problem as follows

- The knowledge base contains the sentences
  "$\%(T, S_1) = [p_1, q_1]$", "$\%(T, S_2) = [p_2, q_2]$",
  "$\%(T, S_3) = [p_3, q_3]$", ...,
  "$\%(T, S_{n-1}) = [p_{n-1}, q_{n-1}]$", "$\%(T, S_n) = [p_n, q_n]$"
- and either contains or entails through subset chaining the sentences
  "$o \in S_1$", "$o \in S_2$", "$o \in S_3$", ..., "$o \in S_{n-1}$", "$o \in S_n$"
- We are interested in finding the probability of
  "$o \in T$"

So $S_1, S_2, S_3, \ldots, S_{n-1}, S_n$ are all candidate reference classes for the query. We are assuming that no other knowledge is available; in particular, knowledge about subset relationships between $S_i$'s is not available. If all the intervals $[p_i, q_i]$ nest within each other, the solution is trivial: the candidate with the narrowest interval would be the answer to the query. If, on the other hand, there are conflicts between the intervals, we cannot establish dominance using Kyburg's original rules [Kyburg, 1983] since we do not have the necessary information about the subset relationships. One could give up and return the interval $[0, 1]$ or resort to constructing various subsets of the cross products of the candidate reference classes. The first method is useless, [3] and variations of the second method admit clear-cut counter-examples [Kyburg, 1991].

---
[3]Though it should be noted that we cannot establish better bounds on the interval by purely set theoretic procedures. It is entirely possible for the probability to be high for two candidate reference classes but very low in their intersection and vice versa. In other words, since we do not know anything about the structures of or the relationships between the candidate reference classes, set

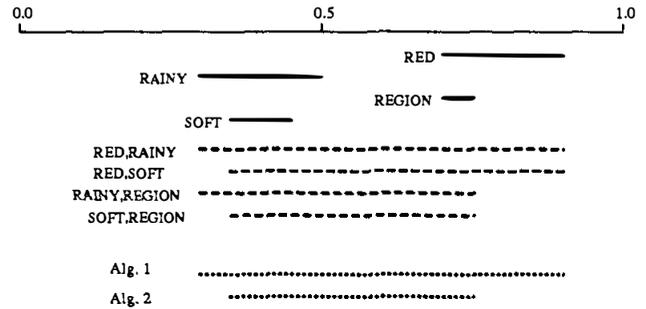

Figure 1: The intervals for the berries example, the dashed lines represent the interval covers of pairs of intervals. The dotted lines are what each algorithm would return. Algorithm 2 is the modified algorithm.

Kyburg used to endorse Loui's approach using subsets of the cross products [Loui, 1986, Kyburg, 1987], but he has changed his mind in recent years [Kyburg, 1991]. He argues that the strongest interval we can justifiably return is the narrowest interval cover that does not conflict with any member of the set of relevant intervals. More formally, given a set of intervals where each member of the set conflicts with at least one other member, we construct an interval using the minimum of the lower endpoints and the maximum of higher endpoints. If there are wide intervals that don't conflict with any other interval in the original set, they can safely be disregarded since they are guaranteed to be as weak or weaker than the narrowest interval cover of the conflicting ones. Thus, in the berries example Kyburg would return the interval $[0.30, 0.90]$ (fig. 1).

## 2  The Proposed Solution

One can think of the procedure proposed in [Kyburg, 1991] as looking at pairs of candidate reference classes and constructing new candidates by taking the interval covers of the conflicting pairs as a means of settling the conflicts.[4] This procedure can be repeated until there is an interval that does not conflict with any of the others. An inefficient but nevertheless illustrative way of computing the interval cover Kyburg would select is given by the following pseudo-code: [5]

---
theory does not help us come up with non-trivial bounds for their intersection which the object belongs to.

[4]Note that we do not have enough data to choose one candidate over another.

[5]Since no information about the subset relationships between the classes is available, we will deal only with the intervals associated with the classes. Even when enough information is available for using Kyburg's rules, it is conceivable that one could end up with a set of candidates rather than a single reference class. So the proposed procedure can be used as the last step of Kyburg's method in the general case.

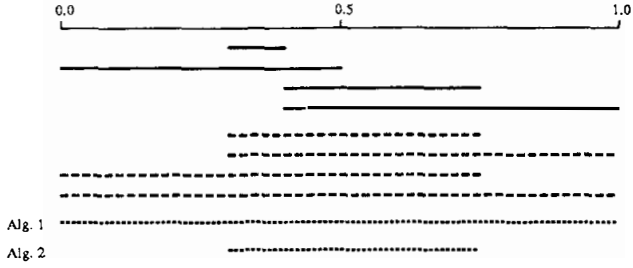

Figure 2: The dashed lines represent the pairwise covers generated in the first iteration of algorithm 1 (elements of $\mathcal{L}'$ after the first iteration). The dotted lines show what each algorithm returns.

### Algorithm 1

```
input: a set L of intervals I_i
repeat
    L' := {}
    for every interval pair I_i = [p_i, q_i], I_j = [p_j, q_j] in L
        if I_i conflicts with I_j then
            I' := [min(p_i, p_j), max(q_i, q_j)]
            L' := L' ∪ {I'}
            mark both I_i and I_j
    L := L ∪ L'
until L' = {}
return the narrowest un-marked interval in L
```

As can be noticed from the pseudo-code, intervals that are no longer candidates (i.e., they are "marked") can still interfere with the selection of other intervals. One upshot of this is that individual "marked" intervals prevent the selection of a narrower cover even when their own cover would not have interfered. This interference from wide intervals is not desirable because it leads us to weaker conclusions. For example, if our set of intervals were $\{[0.3, 0.4], [0.0, 0.5], [0.4, 0.7], [0.4, 1.0]\}$ (Fig. 2), we would have to return $[0.0, 1.0]$, but if we look at the set $\mathcal{L}'$ after the first iteration of the algorithm ($\{[0.3, 0.7], [0.3, 1.0], [0.0, 0.7], [0.0, 1.0]\}$), it is apparent that $[0.3, 0.7]$ is not challenged by any other interval constructed in this iteration. Now, favoring $[0.3, 0.7]$ over the conservative but useless $[0.0, 1.0]$ is appealing because it leads to a stronger result, but can we justify doing so? We can if we are willing to say that interval covers reflect the information represented by their constituents. In the case of two conflicting intervals $[0.0, 0.5]$ and $[0.4, 0.7]$, we might argue that the cover $[0.0, 0.7]$ encodes all we know, given those two bits of information. If we do not go back and look at its constituents, the cover $[0.0, 0.7]$ does not interfere with the stronger cover $[0.3, 0.7]$ even though its constituents ($[0.0, 0.5]$ and $[0.4, 0.7]$) would.

Considering the presence of conflicting evidence, the widening caused by taking covers of intervals is desirable in terms of the semantics one would like to attribute to intervals. Intuitively, one does expect conflicting pieces of evidence to weaken the conclusions, and the interval cover idea nicely captures that intuition. One may not, however, want the weak pieces of evidence to undermine the stronger conclusions indicated by the stronger pieces of evidence with which they are consistent. One way of preventing the weaker evidence from interfering is to disregard or delete the original intervals once we construct all the covers they cause to be constructed. On the other hand, one wants to use all the information available, and actually deleting the constituents once the cover is constructed is not compatible with that ideal. It is not, however, altogether unreasonable to buy into the former argument while keeping the latter in mind. While professing ignorance is a virtue, one should also be able to make the best of available information.

As is the case with Kyburg's method in [Kyburg, 1991], the reference class associated with the interval we return can be any one of the candidate classes associated with the constituents of the cover.

The modified algorithm, which leads to less conservative conclusions, is as follows:

### Algorithm 2

```
input: a set L of intervals I_i
repeat
    L' := {}
    for every interval pair I_i = [p_i, q_i], I_j = [p_j, q_j] in L
        if I_i conflicts with I_j then
            I' := [min(p_i, p_j), max(q_i, q_j)]
            L' := L' ∪ {I'}
            mark both I_i and I_j
    delete all the marked elements of L
    L := L ∪ L'
until L' = {}
return the narrowest un-marked interval in L
```

This algorithm returns the narrowest interval cover whose set of constituents $\mathcal{S} \subseteq \mathcal{L}$ has the property that for any interval $I \in \mathcal{L} - \mathcal{S}$, there is an interval $I^* \in \mathcal{S}$ that nests in and is at least as narrow as $I$. Having made that observation, we can write a more efficient version of the algorithm which illustrates this point.

### Algorithm 2'

```
input: a list L of intervals I_i sorted in ascending order of width
L' := {}
repeat
    extract the first unmarked interval I* from L
    L' := L' ∪ {I*}
    for every remaining un-marked interval I in L
        if I agrees with I* then
            mark I
until there are no more un-marked intervals in L
return interval cover of the intervals in L'
```

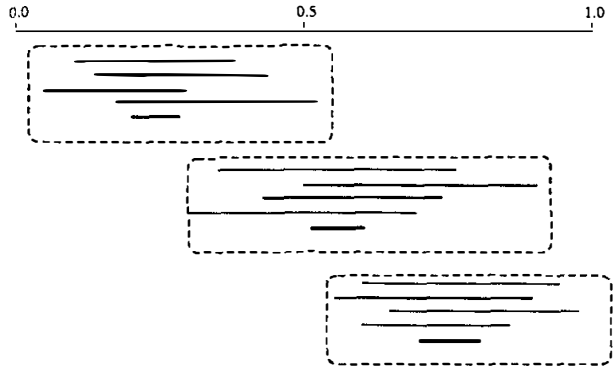

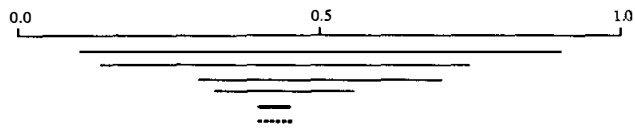

Figure 3: A picture of a more general case with many intervals. The dashed boxes represent sets of intervals with a common narrow sub-interval, which is denoted by a thick line.

Figure 4: The case where there are no conflicts between the candidates.

One can think of each $I^*$ selected in the algorithm as representing the opinion of an independent agent who has access only to the set of intervals $S^* \subseteq \mathcal{L}$ such that for every interval $I \in S^*$, $I^* \subseteq I$ (Fig. 3). The cover of all such $I^*$ would give us the interval on which all such agents would agree.[6] Once again, the point can be made that those agents would not hold those opinions if they had access to what the other agents knew.

### 2.1 Observations

The interval returned by the modified algorithm is never in conflict with the one returned by the original and is at least as narrow. The modified algorithm does return the same result as the original in such cases as the following:

- When there is no conflict between members of the original set of candidate intervals (Fig. 4). This is the most desirable case in that neither algorithm uses evidence combination to obtain a result.

- When there are no two nesting intervals in the original set of candidates (Fig. 5). Having no agreement at all among the pieces of evidence indicates that a very conservative conclusion is called for.

---
[6]Note that we are not requiring all members of $S^*$'s to agree with each other.

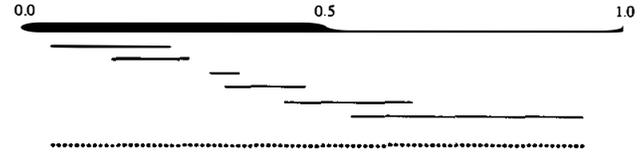

Figure 5: The case where none of the candidates agree.

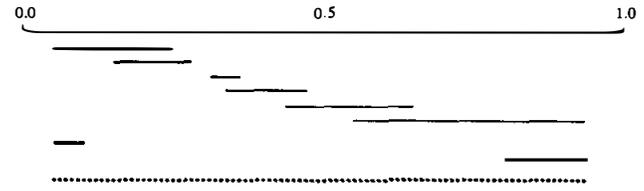

Figure 6: The case where the narrowest intervals (thick lines) are at the extreme ends of the cover of all candidates.

- When the set of narrowest intervals have members that are at the extreme ends of the wider intervals in which they nest (Fig. 6). As can be seen in the figure, having strong but extremely conflicting evidence leads to weak results.

## 3  Conclusions

Avoiding interference from conflicting unreliable data is a problem for autonomous agents except when the designer can hand pick the relevant information. We think playing it safe with large amounts of conflicting data causes pieces of weak evidence to unnecessarily weaken the results of the inference process. The proposed algorithm leads to stronger results by favoring stronger conclusions when there is enough data to justify them.

### Acknowledgments

Research underlying this work has been supported in part by U.S. Army Communication-Electronics Command grant no. DAAB10-87-K-022, and NSF research grant no. IRI-9002659.